\documentclass{article}

\usepackage[preprint]{neurips_2024}
\usepackage{graphicx}
\usepackage{algorithm}
\usepackage{algpseudocode}

\usepackage{booktabs}
\usepackage{tabularx}
\usepackage{subcaption}
\usepackage{amsmath,amssymb}

\usepackage[utf8]{inputenc} 
\usepackage[T1]{fontenc}    
\usepackage{hyperref}       
\usepackage{url}            
\usepackage{booktabs}       
\usepackage{amsfonts}       
\usepackage{nicefrac}       
\usepackage{microtype}      
\usepackage{xcolor}         

\title{Monthly Diffusion v0.9: A Latent Diffusion Model for the First AI-MIP}

\author{
Kyle J. C. Hall\\
\texttt{kylehall@umd.edu} \\
\\
\bf{Maria J. Molina} \\
\texttt{mjmolina@umd.edu}\\
\\
Department of Atmospheric \& Oceanic Sciences\\
  University of Maryland\\
  College Park, MD 20740 \\
}

\begin{document}

\maketitle

\begin{abstract}
Here, we describe Monthly Diffusion at 1.5-degree grid spacing (MD-1.5 version 0.9), a climate emulator that leverages a spherical Fourier neural operator (SFNO)-inspired Conditional Variational Auto-Encoder (CVAE) architecture to model the evolution of low-frequency internal atmospheric variability using latent diffusion. MDv0.9 was designed to forward-step at monthly mean timesteps in a data-sparse regime, using modest computational requirements. This work describes the motivation behind the architecture design, the MDv0.9 training procedure, and initial results. 
\end{abstract}

\section{Introduction}

We trained an artificial intelligence (AI)-based emulator of the monthly mean atmosphere using the ECMWF Reanalysis version 5 (ERA5) \citep{hersbach_era5_2020} with oceanic forcings for the first AI model intercomparison project (AI-MIP). Monthly Diffusion at 1.5-degree grid spacing (MD-1.5 version 0.9) follows the AI-MIP protocol and was created as a pilot architecture for long-timescale climate emulation. Most AI-based atmospheric emulators are trained at short timesteps ($\leq12$-hr) and are optimized for skillful predictions at short lead times ($<$15 days) \citep{brenowitz_climate_2025, cachay_probabilistic_2024, kochkov_neural_2024, lang_aifs_2024, lam2023graphcastlearningskillfulmediumrange, price_probabilistic_2025, pathak_fourcastnet_2022}. At monthly and longer timescales, however, atmospheric predictability is expected to arise primarily from oceanic and coupled processes rather than from atmospheric initial conditions alone.

We use this known predictability to motivate a monthly-timestep emulator that focuses on modes of internal atmospheric variability that evolve slowly enough to be resolved with monthly data, including phenomena such as the North Atlantic Oscillation (NAO). Our model was designed both to emulate internal atmospheric variability over long rollout periods and to assess the atmospheric response to uniform global increases in sea surface temperatures (SSTs). These goals permit substantially lower computational cost than conventional global weather emulators trained at sub-daily timesteps. 

Following the AI-MIP protocol, we submitted 46.25-year historical, $+2$K, and $+4$K SST-forced roll-outs. Here we describe MDv0.9, explain the motivations behind our design choices, recount training protocols, and report initial results. We find that the model can stably emulate the atmosphere for several decades at relatively low GPU cost, although challenges remain in extrapolation beyond the training domain.

\section{Data}

MDv0.9 uses a minimal atmospheric prognostic state, $x_t$, consistent with the AI-MIP protocol, including seven surface or single-level variables (SKT, MTPR, 2D, 2T, SP, 10U, and 10V) and five atmospheric variables (U, V, Q, T, and Z) on seven pressure levels (1000, 850, 700, 500, 250, 100, and 50 hPa). There are also four time-invariant fields, $s$ (ANOR, ISOR, SLOR, and SDOR) (Table \ref{tab:vars}). We use sea surface temperature (SST), sea ice cover (SIC), and land-sea mask as external forcings ($f_t$). The seasonality embedding $e_t^{\mathrm{season}}$ and forcing tensors $f_t$ are concatenated along the channel dimension and treated as a single conditioning tensor $c_t$. Table \ref{tab:quantdesc} further establishes the nomenclature used throughout this work.

\begin{table}[]
    \centering
    \begin{tabular}{c|cc}
        Short Name & Name & Level(s) \\
        \hline
        SKT & skin temperature & surface/single-level\\
        MTPR & mean total precipitation rate & surface/single-level\\
        2D & 2 meter dewpoint temperature & surface/single-level\\
        2T & 2 meter temperature & surface/single-level\\
        SP & surface pressure & surface/single-level\\
        10U & 10 meter U wind component & surface/single-level\\
        10V & 10 meter V wind component & surface/single-level\\
        U & u component of wind & 1000, 850, 700, 500, 250, 100, 50 hPa\\
        V & v component of wind & 1000, 850, 700, 500, 250, 100, 50 hPa\\
        Q & specific humidity & 1000, 850, 700, 500, 250, 100, 50 hPa\\
        T & temperature & 1000, 850, 700, 500, 250, 100, 50 hPa\\
        Z & geopotential & 1000, 850, 700, 500, 250, 100, 50 hPa\\
        ANOR & angle of sub-gridscale orography & invariant (in time)\\
        ISOR & anisotropy of sub-gridscale orography & invariant (in time)\\
        SLOR & slope of sub-gridscale orography & invariant (in time)\\
        SDOR & standard deviation of sub-gridscale orography & invariant (in time)\\
        SST & sea surface temperature & surface/single-level\\
        SIC & sea ice cover & surface/single-level \\
        LSM & land-sea mask & invariant (in time)\\
    \end{tabular}
    \caption{Variables used for MDv0.9.}
    \label{tab:vars}
\end{table}

\begin{table}[]
    \centering
    \begin{tabularx}{\textwidth}{p{0.12\textwidth}|X}
    Quantity & Description \\
    \midrule
    $x_t$ & prognostic atmospheric state, $x_t \in \mathbb{R}^{42 \times 121 \times 240}$ \\
    $f_t$ & external forcing fields (SST, SIC, LSM), $f_t \in \mathbb{R}^{3 \times 121 \times 240}$ \\
    $s$ & static invariant fields (ANOR, ISOR, SLOR, SDOR), $s \in \mathbb{R}^{4 \times 121 \times 240}$ \\
    $m_t$ & $m_t \in [1..12]$ corresponding to months of the year. \\ 
    $e_t^{\mathrm{season}}$ & Contractions of sample-wise seasonal coefficients with learned seasonal basis, $SE(m) \otimes SB = e_t^{\mathrm{season}} \in \mathbb{R}^{3 \times 121 \times 240}$ \\
    $c_t$ & time-varying conditioning tensor, $c_t=[f_t;e_t^{\mathrm{season}}] \in \mathbb{R}^{6 \times 121 \times 240}$ \\
    $\mu_t$ & latent posterior mean, $\mu_t \in \mathbb{R}^{32 \times 40 \times 80}$ \\
    $\sigma_t$ & latent posterior standard deviation, $\sigma_t \in \mathbb{R}^{32 \times 40 \times 80}$. Encoder produces $ln\ \sigma_t^2$, and $\sigma$ is derived by $\exp(0.5\times\log(\sigma^2_t))$. \\
    $z_t$ & sampled latent state, $z_t \in \mathbb{R}^{32 \times 40 \times 80} = \mu_t + \sigma_t\odot\epsilon_v$ where $\epsilon_v \sim \mathcal{N}(0, I)$  \\
    $\epsilon_v$ & I.I.D. Gaussian noise used during the reparametrization trick to sample the latent posterior. \\
    $\epsilon_d$ & I.I.D. Gaussian noise used during reverse denoising \\
    \end{tabularx}
    \caption{State (initial and prognostic) $x_t$, conditioning $c_t$, and latent representation.}
    \label{tab:quantdesc}
\end{table}

All variables were derived from ERA5 monthly means and conservatively (normed) regridded using xESMF \citep{zhuang_pangeo-dataxesmf_2025} to a $1.5\times1.5$-degree equiangular (latitude-longitude) grid, with all variables except MTPR using fractional-area normalization rather than destination-area normalization. Most model development occurred at 5 degrees and 1.5 degrees, and the 1.5-degree output was regridded (conservatively, as above) to a common 1-degree grid for analysis following the AI-MIP protocol. 

We used contiguous training, validation, and test splits, with the training period from January 1, 1985, through December 31, 2014, the validation period from January 1, 1979, through December 31, 1984, and the AI-MIP protocol-specified test period from January 1, 2015, through December 31, 2022. A total of $N=360$ training samples, and $N=72$ validation samples were available to train the emulator with a physical-state input tensor $x_t$ of shape $49C\times121H\times240W$ at 1.5 degrees or $49C\times36H\times72W$ at 5 degrees. 

Prior to training, we standardized each input field by removing a per-variable mean and dividing by a per-variable standard deviation, with specific pressure-level statistics where relevant. Variables that display non-Gaussian distributions were transformed prior to standardization as follows. Non-negative variables (e.g., precipitation and specific humidity) were transformed by taking their square root. We applied a logit transformation to bounded variables (e.g., sea-ice cover (\%) ). After prediction and decoding, transformations and scalings were inverted back to physical units. Reconstruction loss was calculated in transformed, scaled space rather than in physical units.

\section{Architecture}

\begin{figure}[h]
\centering
\includegraphics[width=1\linewidth]{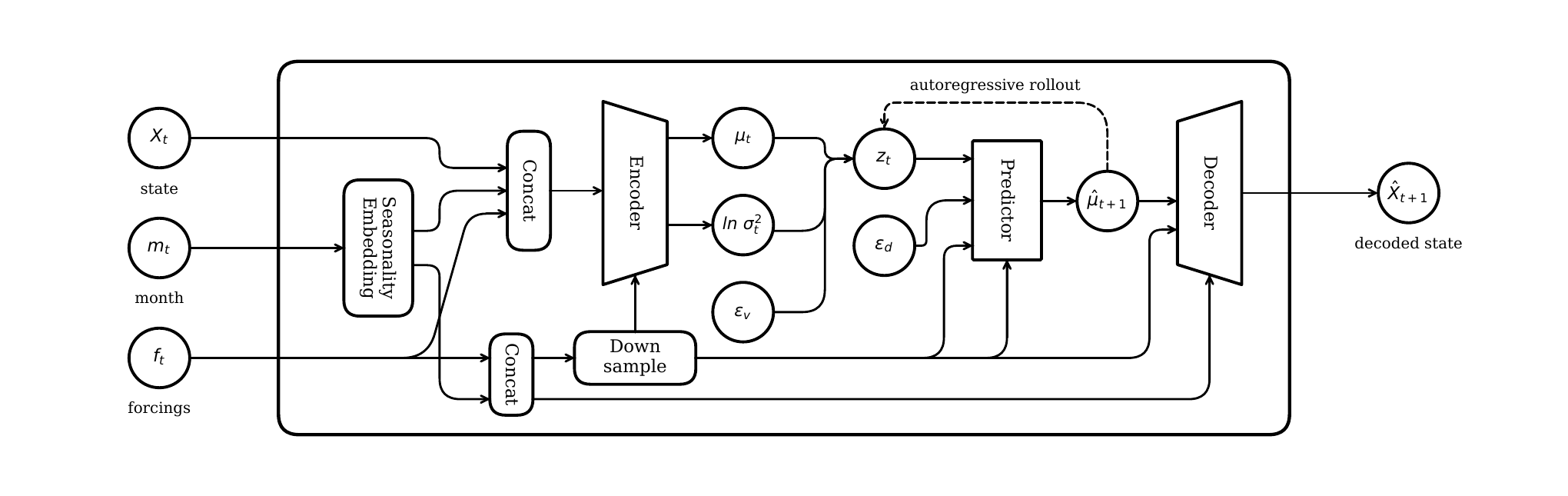}
\caption{Schematic of MDv0.9. Data are denoted by circles, neural networks are denoted by polygons, and subordinate operations are denoted by round-cornered rectangles. Arrows entering laterally indicate data streams used as direct inputs to the indicated operation, and arrows entering from below indicate data streams used as conditioning tensors.}
\label{fig:1}
\end{figure}

MDv0.9 consists of three neural networks, each based loosely on Spherical Fourier Neural Operators (SFNO) \citep{bonev2023sphericalfourierneuraloperators}. Two networks compose the encoder and decoder of a conditional variational autoencoder (CVAE), and the third is a conditional latent diffusion model that predicts the distribution of the next-month latent state (labeled `predictor' in Fig. \ref{fig:1}). During inference, the encoder maps the current atmospheric state $x_t$ to a stochastic latent representation $z_t$, sampled from a latent posterior defined by $\mu_t$ and $\log\sigma_t^2$; the predictor evolves that latent representation $z_t$ forward by one month, producing $\mu_{t+1}$ under conditioning $c_t$; and the decoder maps the predicted latent back to physical space $\hat{x}_{t+1}$ (Fig. \ref{fig:1}).

To clarify notation, let $x_t$ denote the atmospheric state at month $t$, $f_t$ denote the oceanic forcings at month $t$, and $s$ denote the time-invariant static fields. We represent seasonality using a learned embedding $e_t^\mathrm{season}$, and define the full time-varying conditioning vector as $c_t=[f_t,e_t^\mathrm{season}]$.

\subsection{Conditional Variational Autoencoder}

Following previous work on variational autoencoders and conditional variational autoencoders \citep{cvae} \citep{kingma2022autoencodingvariationalbayes, burgess2018understandingdisentanglingbetavae, diederik_introduction_2019}, we model the approximate posterior distribution over latent variables $z_t$ using an encoder network parameterized by $\phi$. Given the atmospheric state $x_t$, seasonality embedding $e_t^\mathrm{season}$, and time-invariant fields $s$, the encoder outputs the latent mean $\mu_t$ and latent log-variance $\log \sigma_t^2$ corresponding to a diagonal-covariance multivariate Gaussian posterior distribution:

$$E_\phi(x_t, c_t, s) = (\mu_t, \log \sigma_t^2),$$

$$q_\phi(z_t|x_t, c_t, s) = \mathcal{N}(\mu_t, \mathrm{diag}(\sigma_t^2)).$$

A latent sample is obtained via the reparameterization trick, $z_t = \mu_t + \sigma_t \odot \epsilon$, where $\epsilon \sim \mathcal{N}(0, I)$,

\noindent which permits gradients to flow through the sampling step during training.

The decoder, parameterized by $\theta$, defines the conditional likelihood of the atmospheric state $x_t$ given the latent sample $z_t$ and the conditioning variables $c_t$:

$$p_\theta(x_t|z_t, c_t).$$

We use a standard isotropic Gaussian prior over the latent variables, 

$$p(z_t) =\mathcal{N}(0, I).$$ 

During training, the CVAE minimizes a loss consisting of a reconstruction term and a Kullback-Leibler regularization term:

$$\mathcal{L_\text{CVAE}} 
= \mathbb{E} \big[||\hat{x}_t-x_t||_2^2 \big] 
+ \lambda_{KL} \,\mathrm{KL} \big( q_\phi(z_t \mid x_t, c_t, s) \, \| \, p(z_t) \big),\qquad \lambda_{KL}\approx0.01,$$

\noindent where $\hat{x}_t$ denotes the decoder reconstruction of $x_t$. The reconstruction term is implemented as mean-squared error, corresponding to a Gaussian likelihood with fixed variance. For a diagonal Gaussian posterior and standard normal prior, the KL term is

\begin{equation}
    \text{KL}(q_\phi(z_t|x_t, c_t, s)\,||\,p(z_t))=\frac{1}{2}\sum_{j=1}^{d_z}(\mu_{t,j}^2+\sigma_{t,j}^2-1-\text{log}\,\sigma_{t,j}^2).
\end{equation}

\noindent Here, $t$ indexes physical monthly time and $j$ indexes latent dimensions. The latent space has shape $32C\times 40H\times 80W$, resulting in a compression ratio of ~0.09 (~11x). However, this ratio is only approximate due to the conditioning of the encoder and decoder.

\subsection{Dual-stream Conditioning}

\begin{figure}[h]
\centering
\includegraphics[width=1\linewidth]{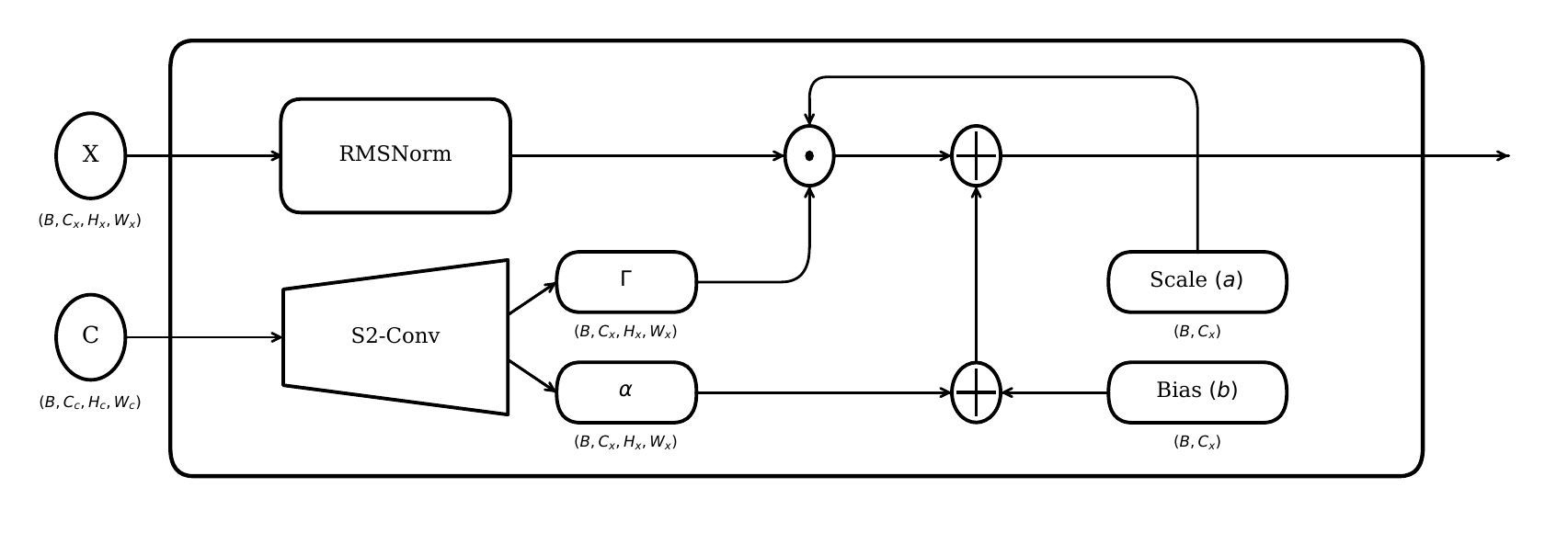}
\caption{Schematic of spatial conditioning RMS normalization layer. }
\label{fig:2}
\end{figure}

We condition the encoder, decoder, and predictor using a dual-stream strategy. The motivation is to allow the latent variables to focus on variability not already explained by known external forcings and the seasonal cycle. For each conditioned network component (encoder, decoder, and predictor), the time-varying conditioning tensor $c_t$ is incorporated in two ways. First, $c_t$ is concatenated with the primary input tensor along the channel dimension. Second, $c_t$ is provided to a spatially conditional normalization layer, which modulates intermediate activations as a function of the conditioning inputs (Fig. \ref{fig:2}). This normalization layer is inspired by Feature-wise Linear Modulation (FiLM) \citep{FiLM}. 

In preliminary experiments, standard normalization layers, including RMSNorm, BatchNorm, and LayerNorm, appeared to attenuate the effect of the conditioning inputs. To address this, we use an RMS normalization followed by a learned affine modulation with spatially varying conditional parameters $\Gamma(c_t)$ and $\alpha(c_t)$. These parameters are predicted from the conditioning tensor $c_t$ using an S2-convolution layer, described below, and are combined with learned channel-wise base parameters $a$ and $b$. 

$$\mathrm{RMSNorm}(h) = \frac{h}{\sqrt(\mathrm{mean}(h^2) + \epsilon)},$$

$$\hat{h} = \mathrm{RMSNorm}(h) \odot \bigl(\Gamma(c_t)+a\bigr) + \bigl(\alpha(c_t)+b\bigr),$$

\noindent where $h$ denotes an intermediate activation tensor. This transformation modulates activations both channel-wise and spatially according to the conditioning inputs.

\begin{figure}[h]
\centering
\includegraphics[width=1\linewidth]{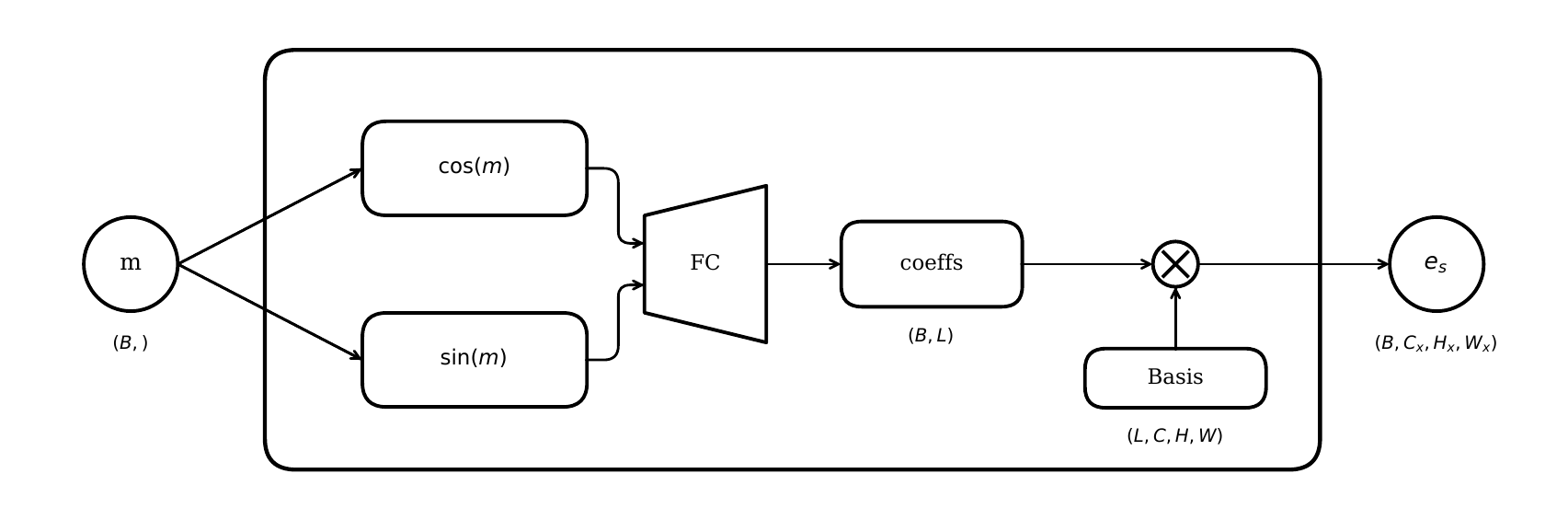}
\caption{Schematic of seasonality embedding.}
\label{fig:3}
\end{figure}

To encourage the latent variables and predictor network to focus on non-seasonal internal variability, we condition all three networks not only on forcings but also on seasonality. Specifically, we embed the month-of-the-year into a learned embedding space using a small fully-connected network. This network predicts the coefficients of a learned basis as a nonlinear function of the annual cycle, yielding a seasonality embedding $e_t^\mathrm{season}$ (Fig. \ref{fig:3}). 


\subsection{Spectral S2-Convolution Layers}

\begin{figure}[h]
\centering
\includegraphics[width=1\linewidth]{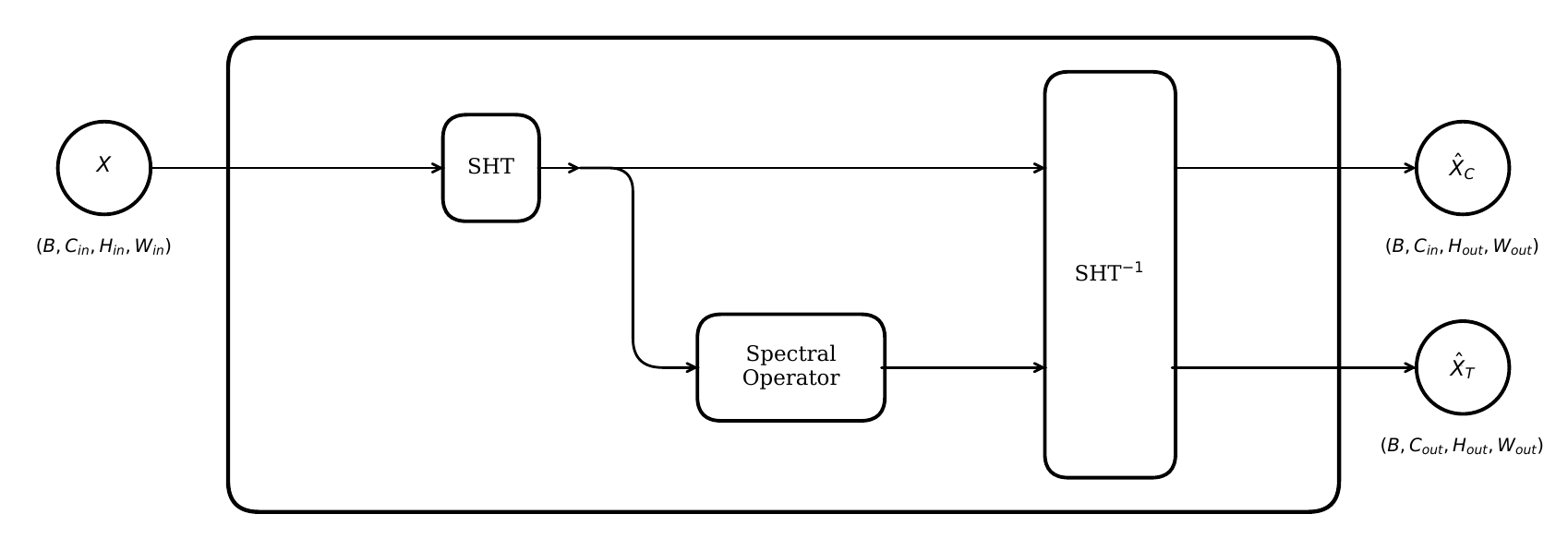}
\caption{Schematic of an S2-convolution layer. A spectral operator is applied to the internal feature tensor after a spherical harmonic transform, after which the result is inverted onto a new spatial grid. }
\label{fig:4}
\end{figure}

The encoder, decoder, and predictor are built from spectral S2-convolution layers that operate by transforming features from the spatial domain to spherical harmonic space, applying either spectral resampling or a learned spectral operator, and then transforming the result back to the spatial domain (Fig. \ref{fig:4}). Spectral S2-convolution layers follow the general framework of spherical Fourier neural operators (SFNOs) \citep{bonev2023sphericalfourierneuraloperators}.

For resolution changes, we perform spectral resampling by truncating or extending the zonal and meridional modes before applying the inverse spherical harmonic transform. This procedure performs spectral upsampling or downsampling on the sphere and is used to change spatial resolution within the network. 

\begin{figure}[h]
\centering
\includegraphics[width=1\linewidth]{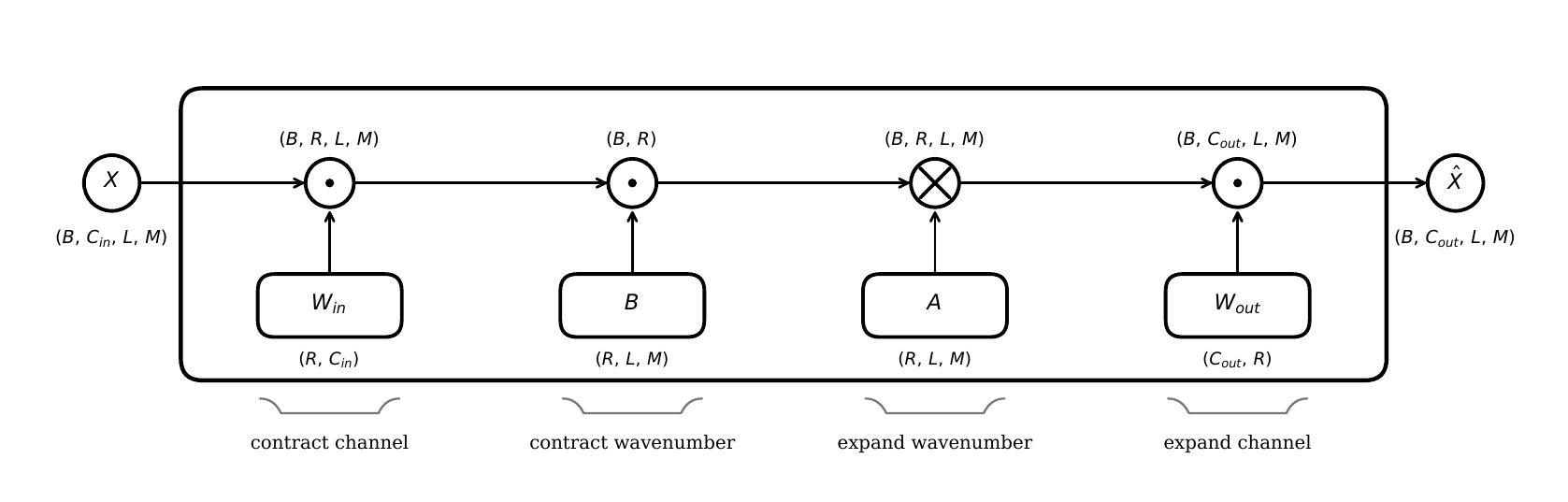}
\caption{Schematic of a tensor-product spectral operator. The shapes of the input tensor and operator parameters are denoted below the corresponding objects, and the dimensionality of the intermediate results is displayed above each operation. Here $B$ is the minibatch dimension, $C_{\mathrm{in}}$ corresponds to the input channels, $L$ and $M$ to the input zonal and meridional wave modes respectively, $R$ to the operator rank, and $C_{\mathrm{out}}$ to the output channels. $\odot$ corresponds to matrix multiplication (inner product) and $\otimes$ to the tensor product (outer product). }
\label{fig:5}
\end{figure}

For learned feature transformations, we apply a trainable spectral operator in spherical harmonic space prior to inversion. Rather than using the standard Driscoll-Healy operator implemented in \texttt{torch-harmonics}, we use a custom low-rank tensor-product spectral operator (Fig. \ref{fig:5}). This operator first projects the channel dimension to an intermediate rank $r$, then contracts the zonal and meridional spectral modes to form a compact spectral representation. The output is subsequently reconstructed through a tensor-product expansion followed by a final channel projection. This factorization couples the channel and spectral structures while using substantially fewer parameters than a dense spectral operator.

\newpage 

\subsection{Encoder, Decoder, and Predictor Networks}

\begin{figure}[h]
\centering
\includegraphics[width=1\linewidth]{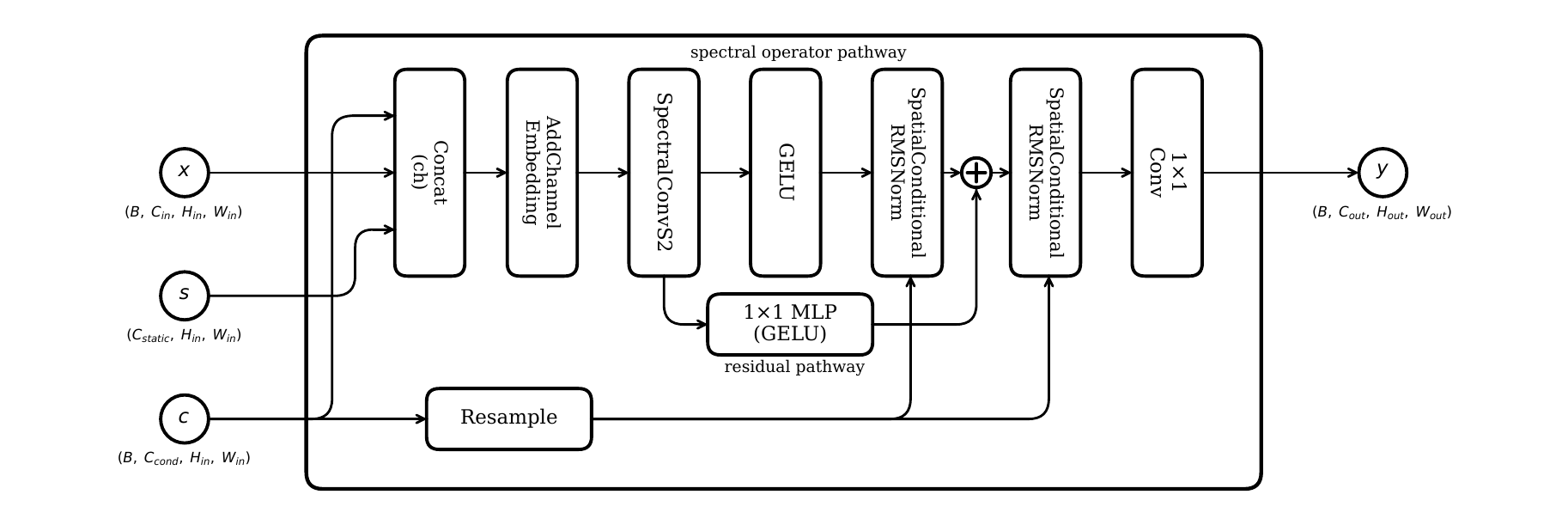}
\caption{Schematic of the encoder. In Fig. \ref{fig:1}, the resampling block in the conditioning pathway is placed outside of the encoder for visual clarity. The residual pathways are omitted from the S2-convolution layers located inside the SpatialConditionalRMSNorm blocks. The decoder uses this same architecture, but with inverted dimensions where appropriate. The predictor uses the same architecture, but with no static fields, and an additional input for DDPM-style noise input.  }
\label{fig:6}
\end{figure}

The encoder, decoder, and predictor networks are all constructed from the spectral S2-convolution and conditional normalization components described above. Although the three networks serve different roles, they share a common architectural pattern with modifications appropriate to their respective inputs and outputs (Fig. \ref{fig:6}). Each network uses a single spectral-operator-based compressive residual block, as shown in Fig. \ref{fig:6}. Tables \ref{tab:mdv09_architecture2} and \ref{tab:mdv09_architecture3} describe architectural details of the encoder-decoder pair and latent prediction networks, respectively. 

\begin{table*}[t]
\centering
\begin{tabularx}{\linewidth}{@{}lX@{}}
\toprule
\textbf{Component} & \textbf{Description} \\
\midrule
Encoder $E_\phi$ & maps $(x_t,c_t,s)$ to posterior parameters $(\mu_t,\log \sigma_t^2)$ where $q_\phi(z_t\mid x_t,c_t,s)=\mathcal{N}(\mu_t,\mathrm{diag}(\sigma_t^2))$ \\
Decoder $D_\theta$ & maps latent state and conditioning back to physical space, modeling $p_\theta(x_t\mid z_t,c_t)$ \\
Seasonal Embedding $SE(m)$ & MLP taking inputs  $[\sin(2\pi\, m/12),\cos(2\pi\, m/12)]$, and producing coefficients for a learned latent seasonal basis tensor $SB$ with shapes: input $ \ (B, 2) \rightarrow $ hidden $(B, 64) \rightarrow $ output $(B,3)$ \\
Learned Seasonal Basis $SB$ & tensor of shape $(3,3,121,240)$. Seasonal embedding derived by the tensor product $SE(m) \otimes SB$, contracting the basis dimension. \\
Shared spectral operator family & low-rank tensor-product spectral operator with channel mixing $W_{\mathrm{in}}$, $W_{\mathrm{out}}$ and spectral factors $A,B$ \\
Encoder spectral rank $r_E$ & 64 \\
Decoder spectral rank $r_D$ & 256 \\
Encoder/Decoder residual MLP width & $1{\times}1$ channel mixer with hidden width 128 \\
Encoder/Decoder cond.-norm rank & 2 / 4, respectively \\
Encoder cond.-norm hidden width & 4 channels in the $\Gamma/\alpha$ subnetwork \\
Decoder cond.-norm hidden width & 2 channels in the $\Gamma/\alpha$ subnetwork \\
Activation & GELU \\
Variable/level metadata embedding & learned channel embedding with embedding dimension 8 \\
Encoder parameters & 1,968,505 trainable parameters at $1.5^\circ$ \\
Decoder parameters & 1,810,487 trainable parameters at $1.5^\circ$ \\
Dual-stream conditioning & conditioning enters through both concatenation and conditional normalization \\
\bottomrule
\end{tabularx}
\caption{Architecture summary for MDv0.9 at $1.5^\circ$ resolution: encoder, decoder, and spectral building blocks.}
\label{tab:mdv09_architecture2}
\end{table*}

\begin{table*}[t]
\centering
\begin{tabularx}{\linewidth}{@{}lX@{}}
\toprule
\textbf{Component} & \textbf{Description} \\
\midrule
Predictor $v_\psi$ & conditional latent diffusion model for the next-month latent target \\
Training target & next-month posterior mean $\mu_{t+1}$, conditioned on stochastic $z_t$ \\
Diffusion steps & $T=15$ \\
Noise schedule & cosine $\beta$-schedule with offset $s=0.008$ \\
Time encoding & sinusoidal diffusion-time embedding followed by MLP \\
Time-embedding width & 32-dimensional sinusoidal embedding projected to 5 spatial channels \\
Predictor spectral rank $r_P$ & 128 \\
Predictor residual MLP width & $1{\times}1$ channel mixer with hidden width 128 \\
Predictor cond.-norm rank & 2 \\
Predictor cond.-norm hidden width & 4 channels in the $\Gamma/\alpha$ subnetwork \\
Latent normalization & learned scalar centering and scaling parameters $(\mu_p,\sigma_p)$ \\
Predictor parameters & 951,507 trainable parameters at $1.5^\circ$ \\
\bottomrule
\end{tabularx}
\caption{Architecture summary for MDv0.9 at $1.5^\circ$ resolution: latent diffusion predictor and conditioning pathway.}\label{tab:mdv09_architecture3}
\end{table*}

The encoder maps the current atmospheric state $x_t$, together with the time-varying conditioning vector $c_t$ and static fields $s$, to the parameters of the variational posterior distribution over latent variables. Specifically, it produces the latent mean $\mu_t$ and latent log variance $\log\sigma_t^2$, from which a latent sample $z_t$ is drawn via the reparameterization trick. The encoder has 1,968,505 trainable parameters at 1.5 degrees.

The predictor operates in latent space. Conditioned on the current latent state $z_t$, the time-varying conditioning vector $c_t$, and the diffusion-time embedding, it predicts the denoising target required to generate the next latent state $\hat{\mu}_{t+1}$. During autoregressive rollout, the predicted latent state $\hat{\mu}_{t+1}$ is fed back into the predictor as the latent input for the following step. The predictor consists of 951,507 trainable parameters at 1.5 degrees, given a compression ratio of ~0.09.  

The decoder maps a latent state and conditioning information back to physical space. During training, it reconstructs the current atmospheric state from $z_t$ and $c_t$. During autoregressive prediction, it maps the predicted next latent state $\hat{\mu}_{t+1}$, together with conditioning information for the target month, to the predicted atmospheric state $\hat{x}_{t+1}$. The decoder has 1,810,487 trainable parameters at 1.5 degrees. We use the GELU activation function where applicable.

\subsection{Latent Diffusion with Forced and Stochastic Conditioning}

The predictor network is a conditional latent diffusion model \citep{DBLP:journals/corr/abs-2112-10752, DBLP:journals/corr/abs-2006-11239, DBLP:journals/corr/abs-2102-09672}. It is trained to learn the distribution of the next-month latent mean $\mu_{t+1}$, conditioned on the current stochastic latent sample $z_t$ and the current (downsampled) conditioning tensor $c_t$ \citep{salimans2022progressive}. We therefore model 

$$p_\psi(\mu_{t+1}|z_t, c_t),$$

\noindent where $\psi$ denotes the parameters of the predictor network.

The predictor is trained using $v$-prediction with a cosine $\beta$-schedule and 15 diffusion noise levels. Let $y_0$ denote the normalized target latent mean, $\mathrm{Norm}(\mu_{t+1})$, for month $t+1$. A noised latent $y_k$ at diffusion step $k$ is constructed from the clean latent $y_0$ as 

$$y_k = \sqrt{\bar{\alpha}_k}\, y_0 + \sqrt{1 - \bar{\alpha}_k}\,\epsilon, \qquad
\epsilon \sim \mathcal{N}(0, I),$$

\noindent where $k$ indexes diffusion time and $\bar{\alpha}_k$ is the cumulative noise schedule. The predictor receives the normalized current latent $\mathrm{Norm}(z_{t})$, the noised target latent $y_k$, the conditioning vector $c_t$, and the normalized diffusion-time index $k/T$, and predicts the $v$-target used for denoising.

The CVAE and predictor are optimized jointly as discussed in Section \ref{jsect}.

\subsubsection{Stochastic Prior-State Conditioning}

We condition the predictor on the stochastic latent sample $z_t$ rather than on the posterior mean $\mu_t$. Empirically, this broadens the support of the conditioning distribution seen during training and reduces sensitivity to small latent errors during autoregressive rollout. We also train the predictor to model the distribution of $\mu_{t+1}$ rather than the distribution of a stochastic sample $z_{t+1}$. This choice reduces the tendency of the transition model to accumulate additional sampling noise during autoregression. Intuitively, the stochasticity of $z_t$ acts as a form of data augmentation in the latent transition task, while predicting the next posterior mean provides a more stable target for the subsequent decoder. 

\subsubsection{Joint Optimization}\label{jsect}

Many latent diffusion models are trained in two stages: first by fitting an autoencoder, and then by fitting a diffusion model on the resulting latent space \citep{DBLP:journals/corr/abs-2112-10752}. Here, we instead jointly train the encoder, decoder, and predictor. Our motivation is that, in a highly data-limited setting, the encoder should learn not only a compact representation of the atmospheric state but also a latent geometry amenable to smooth conditional diffusion dynamics. Importantly, our diffusion network learns to center and scale the latents online, so that they are compatible with i.i.d. DDPM noise. This is conceptually similar to stable diffusion \citep{DBLP:journals/corr/abs-2112-10752}; however, since stable diffusion uses pretrained autoencoders, the latent means and scales can be precomputed offline. 

Our full training objective is then formulated as

\begin{align}
\mathcal{L}_{\mathrm{total}}
&=
\lambda_{\mathrm{rec}}\,\mathcal{L}_{\mathrm{rec}}
+
\lambda_{\mathrm{diff}}\,\mathcal{L}_{\mathrm{diff}}
+
\lambda_{\mathrm{KL}}\,\mathcal{L}_{\mathrm{KL}}
+
\lambda_{\mathrm{std}}\,\mathcal{L}_{\mathrm{std}}
+
\lambda_{\mathrm{mean}}\,\mathcal{L}_{\mathrm{mean}},
\\[6pt]
\mathcal{L}_{\mathrm{rec}}
&=
\mathbb{E}\!\left[
\left\|
\hat{x}_{t}-x_{t}
\right\|_2^2
\right],
\\[6pt]
\mathcal{L}_{\mathrm{diff}}
&=
\mathbb{E}\!\left[
\left\|
\hat{v}_{t+1,k}-v_{t+1,k}
\right\|_2^2
\right],
\\[6pt]
\mathcal{L}_{\mathrm{KL}}
&=
\mathbb{E}\!\left[\mathrm{KL}\!\left(
q_{\phi}(z_{t}\mid x_{t},c_{t},s)
\,\|\, 
p(z_{t})
\right)\right],
\\[6pt]
\mathcal{L}_{\mathrm{std}}
&=
\mathbb{E}\!\left[
\left\|
\bar{\sigma}_{p}-\bar{\sigma}_{t}
\right\|_2^2
\right],
\\[6pt]
\mathcal{L}_{\mathrm{mean}}
&=
\mathbb{E}\!\left[
\left\|
\bar{\mu}_{p}-\bar{\mu}_{t}
\right\|_2^2
\right].
\end{align}

where $\hat{x}_t$ is the reconstructed atmospheric state at time $t$, and $\hat{v}_{t+1, k}$ is the predicted $v$-target at diffusion step $k$ for the latent transition from month $t$ to month $t+1$. Under $v$-prediction, the target is defined as 

$$v = \sqrt{\bar{\alpha}_k}\,\epsilon -\sqrt{1 - \bar{\alpha}_k}\,y_0,$$

\noindent where $y_0$ is the normalized clean latent target and $\epsilon \sim \mathcal{N}(0, I)$. The quantities $\mu_t$ and $\sigma_t$ correspond to the variational latent mean and standard deviation at month $t$. The parameters $\mu_p$ and $\sigma_p$ are the learned scalar normalization parameters used by the denoising network to center and scale latent variables online before diffusion. This architecture also uses positional encodings to represent variable identity and relative position along the vertical dimension, which would otherwise be lost when different pressure levels and variables are treated simply as channels. In addition, a sinusoidal positional encoding is used to represent the diffusion time during denoising. Algorithm 1 summarizes the $v$-prediction and DDPM sampling procedures \citep{DBLP:journals/corr/abs-2006-11239, DBLP:journals/corr/abs-2102-09672, DBLP:journals/corr/abs-2112-10752}. 


\begin{algorithm}[t]
\centering
\begin{minipage}{0.98\linewidth}
\begin{algorithmic}[1]
\Statex \textbf{Algorithm: Conditional latent DDPM with $v$-prediction}
\Statex \textbf{Indices:} physical time $t$; diffusion time $k$
\Statex \textbf{Given:} latent pair $(z_{t},\mu_{t+1})$, conditioning tensor $c_{t}$; diffusion horizon $T$
\Statex \textbf{Given:} noise schedule $\{\beta_k\}_{k=1}^T$, with $\alpha_k=1-\beta_k$ and $\bar{\alpha}_k=\prod_{i=1}^k\alpha_i$
\Statex \textbf{Given:} DDPM posterior variance 
\[
\tilde{\beta}_k = \frac{1-\bar{\alpha}_{k-1}}{1-\bar{\alpha}_k}\,\beta_k
\]
\Statex \textbf{Given:} learned latent normalization parameters $(\mu_p,\sigma_p)$ with
\[
\mathrm{Norm}(z)=\frac{z-\mu_p}{\sigma_p},
\qquad
\mathrm{UnNorm}(z)=z\,\sigma_p+\mu_p
\]
\vspace{2pt}
\Statex

\Statex \textbf{Training (learn $p_{\psi}(\mu_{t+1}\mid z_t,c_t)$ via auxiliary diffusion time)}
\State $y_0 \gets \mathrm{Norm}(\mu_{t+1})$
\State Sample $k \sim \mathrm{Unif}\{1,\dots,T\}$, \quad $\epsilon \sim \mathcal{N}(0,I)$
\State Forward noising: 
\[
y_{k} \gets \sqrt{\bar{\alpha}_k}\,y_0 + \sqrt{1-\bar{\alpha}_k}\,\epsilon
\]
\State $v$-target: 
\[
v_k \gets \sqrt{\bar{\alpha}_k}\,\epsilon - \sqrt{1-\bar{\alpha}_k}\,y_0
\]
\State Predict: 
\[
\hat{v}_k \gets v_\psi\!\big(\mathrm{Norm}(z_t),\,y_{k},\,c_t,\,k/T\big)
\]
\State Minimize denoising loss: 
\[
\mathcal{L}_{\mathrm{diff}} \gets \|\hat{v}_k - v_k\|_2^2
\]
\vspace{3pt}
\Statex

\Statex \textbf{Sampling (one physical step $t \to t+1$)}
\State Initialize $y_{T} \sim \mathcal{N}(0,I)$
\For{$k=T,\dots,1$}
  \State Predict:
  \[
  \hat{v}_k \gets v_\psi\!\big(\mathrm{Norm}(z_t),\,y_k,\,c_t,\,k/T\big)
  \]
  \State Convert $v$-prediction to a noise estimate $\hat{\epsilon}_k$: 
  \[
  \hat{\epsilon}_k \gets \sqrt{\bar\alpha_k}\,\hat v_k + \sqrt{1-\bar\alpha_k}\,y_{k}
  \]
  \State DDPM posterior mean: 
  \[
  \mu_{y,k} \gets \frac{1}{\sqrt{\alpha_k}}\!\left(y_{k} - \frac{\beta_k}{\sqrt{1-\bar{\alpha}_k}}\,\hat{\epsilon}_k\right)
  \]
  \If{$k>1$}
    \State Sample $\epsilon \sim \mathcal{N}(0,I)$
    \State 
    \[
    y_{k-1} \gets \mu_{y,k} + \sqrt{\tilde{\beta}_k}\,\epsilon
    \]
  \Else
    \State 
    \[
    y_{0} \gets \mu_{y,1}
    \]
  \EndIf
\EndFor
\State \textbf{return} 
\[
\hat \mu_{t+1} \gets \mathrm{UnNorm}(y_{0})
\]
\end{algorithmic}
\end{minipage}
\caption{Conditional latent DDPM with $v$-prediction. Physical time $t$ indexes the monthly evolution, while diffusion time $k$ indexes the internal denoising chain used to sample the conditional transition.}
\end{algorithm}

\subsection{Training}

We train the model using the AdamW optimizer with an initial learning rate of 0.001 and weight decay of 0.0001. We use loss weights $\lambda_{rec}=1$, $\lambda_{diff}\approx0.5$, $\lambda_{KL}\approx0.01$, and $\lambda_{std}=\lambda_{mean}=1$. We use a batch size of 4, and models are trained for 100 epochs using best-model check-pointing. At $1.5\times1.5$-grid spacing on an NVIDIA A100 GPU, a typical training run takes approximately 30 minutes. We also maintain an exponential moving average of the model weights with a decay rate of 0.995 to stabilize training.

\section{Forced Ensemble Experiments}

We performed several ensemble experiments to assess MDv0.9's mean climatology, SST-forced responses, and atmospheric internal variability. Each ensemble experiment was initialized in 1978-10 using ERA5 initial conditions, and run autoregressively until 2024-12 (46.25 years). No perturbed initial conditions were used; ensemble stochasticity was introduced purely through latent diffusion stochastic sampling. We did not allow our model sufficient ``spin-up" time, and stratospheric biases (described below) suggest that a spin-up period of at least a decade may be necessary for the internal state to adjust from initial conditions. Nevertheless, we follow the AI-MIP protocol, which was designed primarily for weather-scale models. 

The monthly time scale and low computational cost of MDv0.9 allow auto-regressive ensemble rollouts to be performed quickly. Vectorizing across ensemble members enabled us to run a 46.25-year, 50-member experiment in about twenty minutes. However, analysis quickly becomes unwieldy with very large ensembles, so we opted to use twenty-member ensembles here to balance memory and storage requirements with fair characterization of model stochasticity. 

Our ``historical" run used unmodified ERA5 SST and SIC as external forcings. Our ``historical +2K" run used the same external forcings, with a spatially uniform two Kelvin increase applied at all time steps. We also ran a 46.25-year ``climatology-forced" experiment, where we replaced historical ERA5 SST and SIC with the 1979-2024 climatological average SSTs and SICs, in order to control for interannual and slower variability introduced by the ocean. Finally, we ran a ``climatology +2K" experiment, which added a globally uniform 2K increase to the SST climatology. We did not change the SIC field in either ``+2K" experiment, although this should be revisited in future work for physical consistency. 

\begin{figure}[h]
\centering
\includegraphics[width=1\linewidth]{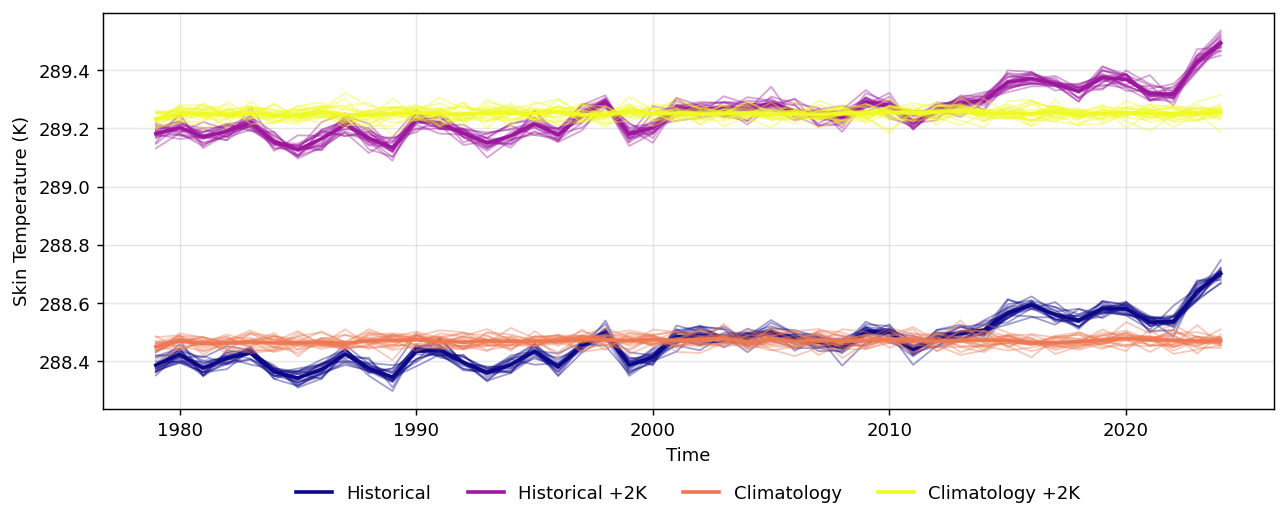}
\caption{Cosine-latitude-weighted annual-mean skin temperature from each of the four sea-surface temperature forced ensemble experiments.}
\label{fig:7}
\end{figure}

Figure \ref{fig:7} displays the annual-average, area-weighted global mean skin temperature over time from each of the four ensemble experiments. The historical and historical +2K experiments display a secular trend and decadal and interannual variability associated with oceanic forcings, while the climatology and climatology +2K experiments do not. Additionally, the +2K experiments display a marked increase of approximately 0.8K in global-mean, annual-average skin temperature. This demonstrates that our model responds to the imposed SST forcings with the correct sign and a reasonable magnitude.  

\begin{figure}[h]
\centering
\includegraphics[width=1\linewidth]{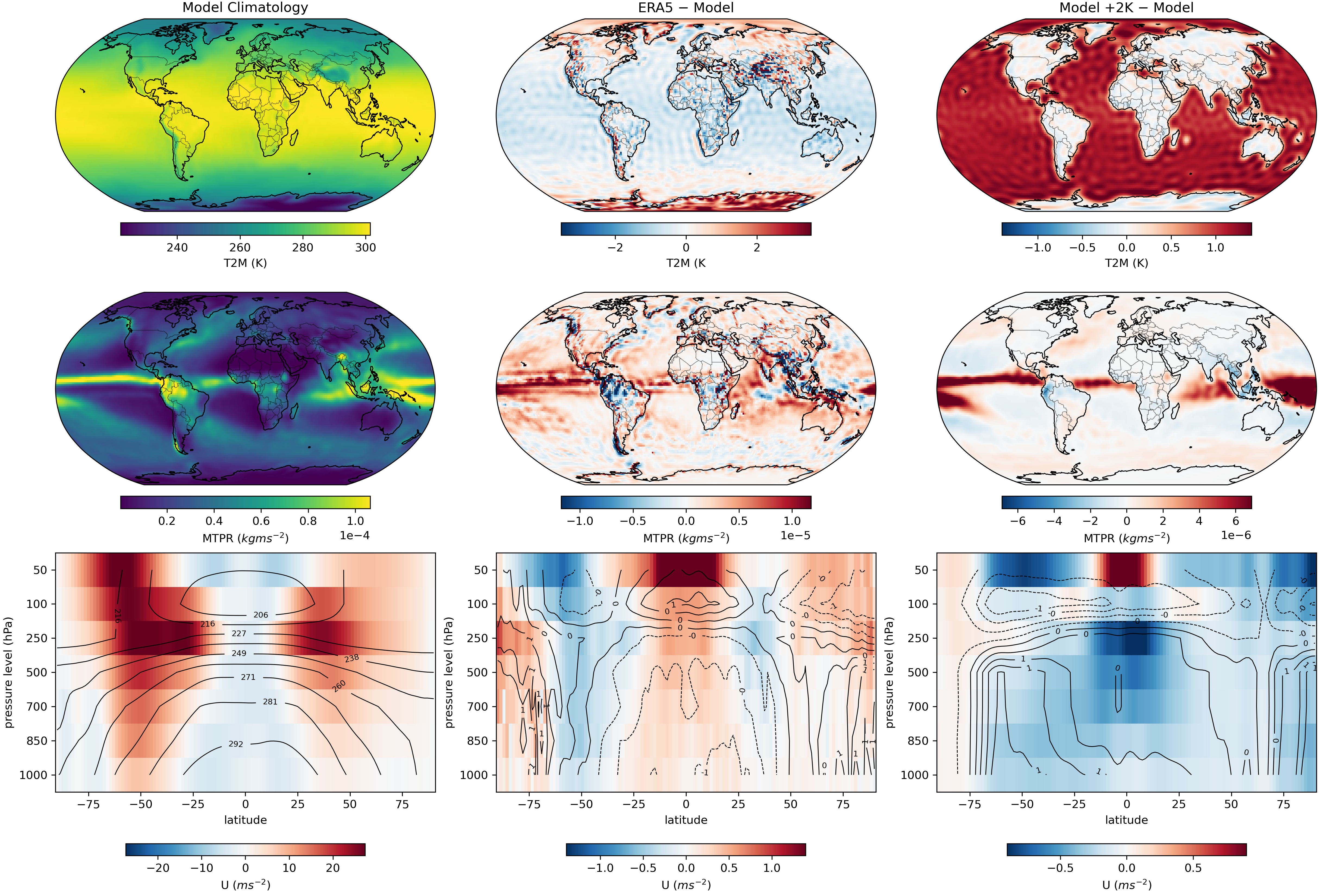}
\caption{Annual-mean climatology of the MDv0.9 historical ensemble mean, differences between ERA5 and MDv0.9 historical ensemble mean, and differences between MDv0.9 historical +2K and MDv0.9 historical ensemble means.}
\label{fig:8}
\end{figure}

\subsection{Biases in Atmospheric Circulation, Temperature, and Precipitation}

Figure \ref{fig:8} displays the annual-mean climatology of the MDv0.9 historical ensemble mean, the differences between the ERA5 annual climatology and the MDv0.9 historical ensemble mean, and the differences between the MDv0.9 historical +2K run and the MDv0.9 historical run. The first row displays global maps of annual average 2-meter temperature in Kelvin; the second, global maps of annual average monthly-mean total precipitation rate (MTPR) in $kg\ m^{-2}s^{-1}$; and the third, zonal-mean zonal wind (shading) and temperature (contours) in the latitude-height plane. 

MDv0.9 displays reasonable climatological spatial distributions of temperature and precipitation, reproducing notable features such as meridional temperature gradients, the intertropical convergence zone (ITCZ), and mid-latitude storm tracks. However, MDv0.9 appears to produce an overly strong climatological meridional temperature gradient during the historical period, as indicated by the warm anomalies at high latitudes and cool anomalies in the tropics (Fig. \ref{fig:8}b). This is consistent with its climatological bias in the vertical structure of the atmosphere, including enhanced midlatitude jets and polar vortex compared to ERA5. Generally, thermal wind balance appears to be broadly maintained, with enhanced upper-level zonal winds corresponding approximately to zonal temperature gradients. 

MDv0.9 generally exhibits less precipitation in the ITCZ than ERA5, and enhanced precipitation over northwest South America and Southeast Asia. MDv0.9's spectral architecture does seem to imprint on the residuals: a small-scale wave pattern is ubiquitously present in the spatial distributions of these variables. The most notable failures of MDv0.9 are over mountainous regions globally, where the model seems to struggle to resolve the small-scale variability associated with topography. One possible explanation is that the decoder does not receive the invariant fields, which should be addressed in future model versions. 

The largest bias in the vertical structure of the atmosphere in MDv0.9 is in the equatorial stratosphere (50 hPa). Stratospheric zonal wind is notably stronger in ERA5 than in the MDv0.9 historical run. This could be related to stratospheric adjustment (``model shock"): we evaluated the quasibiennial oscillation (QBO) in our model compared to that of ERA5 and found that it displayed a large-magnitude bias for approximately the first five years of integration, which eventually subsides and becomes substantially less structured. MDv0.9 is generally colder than ERA5 in the tropical stratosphere and warmer in the tropical/midlatitude troposphere, which could indicate an enhanced Hadley circulation associated with stronger meridional temperature gradients. 

\subsection{Response to +2K SST Forcing}

To assess the response of our model to the uniform 2K global increase in SST, we compare the annual-mean ensemble-mean climatologies of the MDv0.9 historical and +2K ensemble means (Fig. \ref{fig:8}c-i). The global response of precipitation appears to reflect a ``wet-gets-wetter'' paradigm, except over land, which is reversed. Tropical ITCZ and midlatitude storm-track rainfall generally appear to intensify, but precipitation over northwest South America and Southeast Asia appears to decrease. 

Unsurprisingly, 2-meter temperature over the global oceans appears to increase, but no systematic changes occur over the continents (changes are mostly attributable to model-architecture spectral noise). MDv0.9 +2K generally displays a warming troposphere and a cooling stratosphere. Stratospheric cooling is an expected result of increasing CO$_2$ concentrations. Although MDv0.9 does not involve atmospheric composition or radiative forcing, it appears to have learned an association between warmer SST states and this vertical temperature response over the historical period. The Antarctic troposphere also appears to cool, and the westerly zonal wind in the stratosphere intensifies.  

\subsection{Ensemble-Mean Atmospheric Response to ENSO}

\begin{figure}[h]
\centering
\includegraphics[width=1\linewidth]{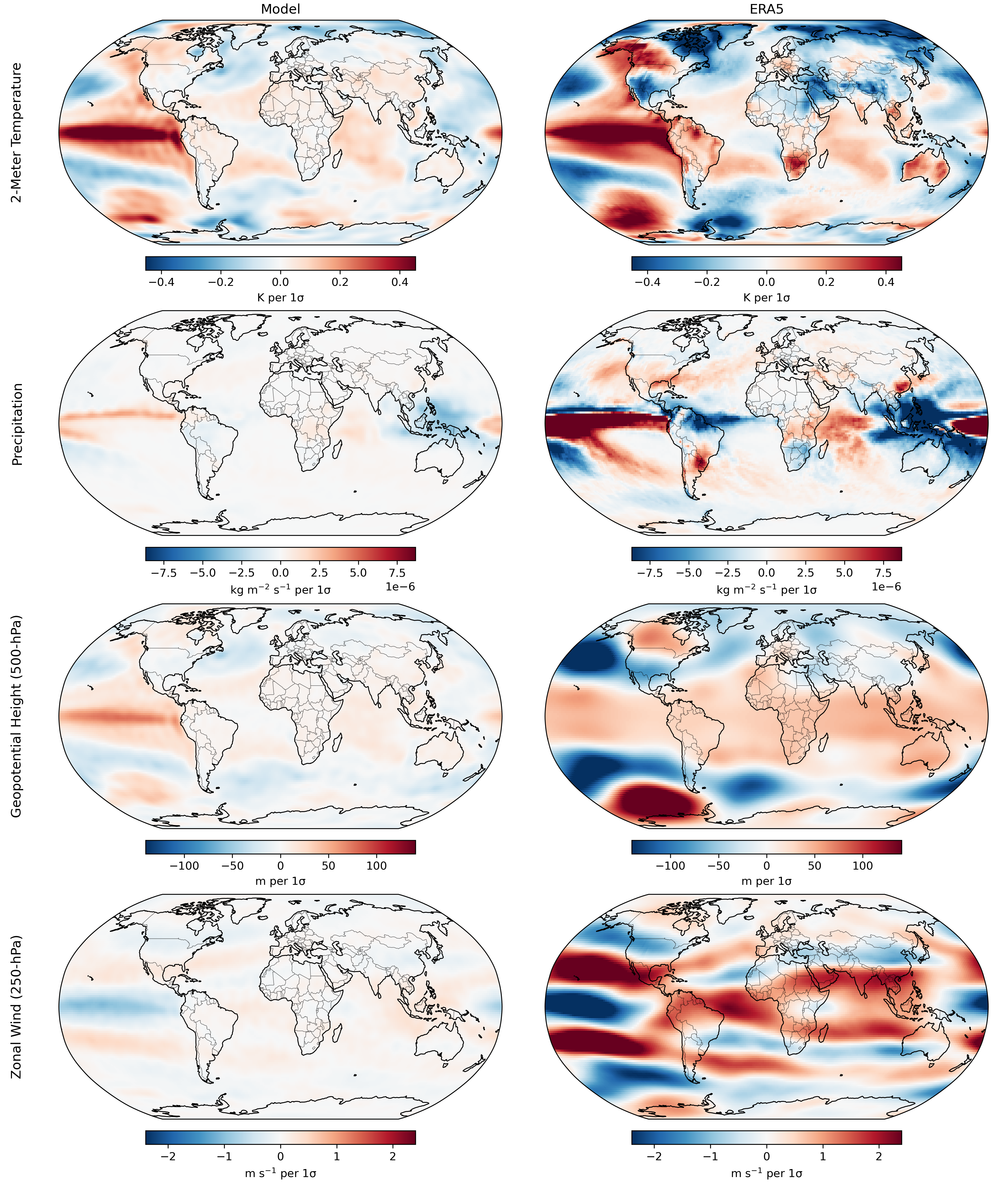}
\caption{ERA5 vs MDv0.9-historical regression on forced Niño-3.4}
\label{fig:9}
\end{figure}

We regressed the ensemble-mean MDv0.9 historical run on a Niño 3.4 index calculated from the historical forcing SST data. We expected that an atmospheric response to ENSO would be shared across ensemble members, since they all see the same SST forcings. Figure \ref{fig:9} displays regressions of the ensemble mean T2M, MTPR, Z500, and U250 anomaly fields on this forced ENSO index, for both the ERA5 reference data and the MDv0.9 historical run. Generally, MDv0.9 displays reasonable ENSO teleconnection patterns in all variables over the global oceans. MTPR and T2M exhibit dramatically reduced teleconnections across continents. Z500 shows an SST-like pattern that is not present in ERA5. In general, the magnitudes of the ENSO teleconnections in MDv0.9 are meaningfully decreased compared to those in ERA5.

\subsection{Atmospheric Internal Variability (North Atlantic Oscillation)}

Performing time-stepping in the latent space of a forcing-conditioned CVAE should allow MDv0.9 to resolve atmospheric internal variability. The North Atlantic Oscillation (NAO) is a notable mode of atmospheric variability. To evaluate NAO representation, we fit the first EOF of Z500 anomalies from ERA5 (0-90N) and projected per-member MDv0.9 climatology-forced Z500 anomalies onto it to quantify NAO-like variability in each member. 

\begin{figure}[h]
\centering
\includegraphics[width=1\linewidth]{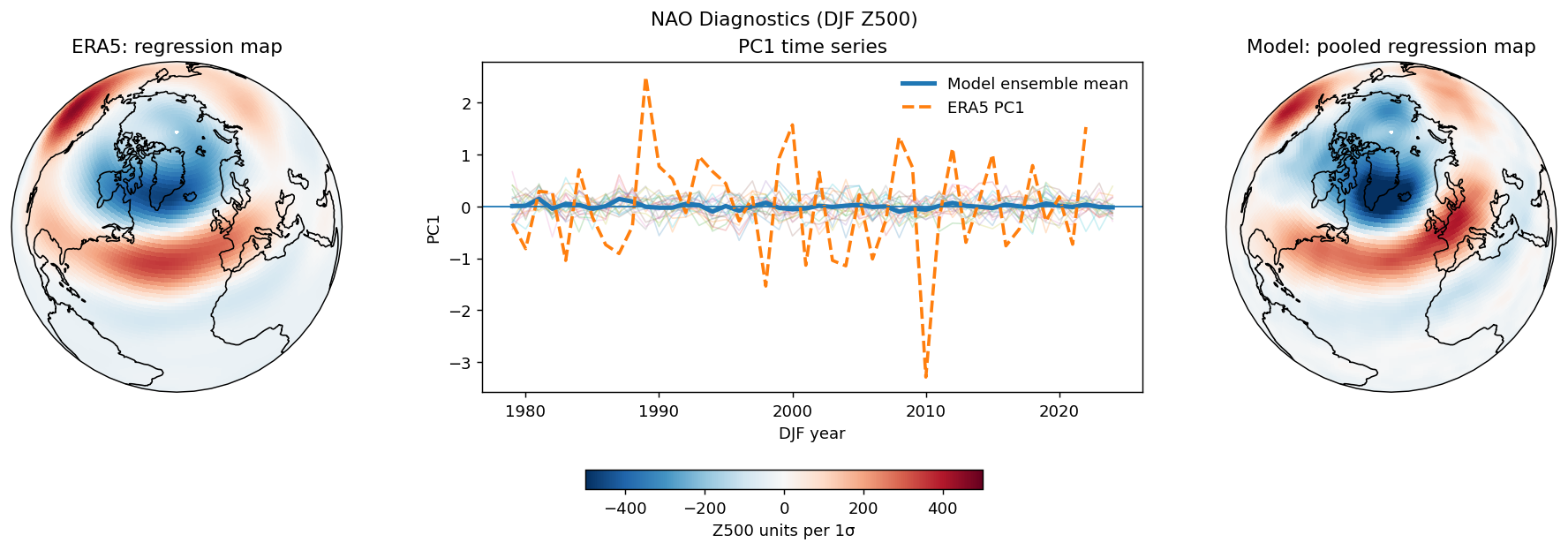}
\caption{Regression onto the leading EOF of Z500 over 0–90°N for ERA5 and the climatology-forced MDv0.9 run.}
\label{fig:10}
\end{figure}

Figure \ref{fig:10}b displays the ERA5 PC1 associated with NAO, as well as the per-member MDv0.9 climatology forced NAO time series based on the same ERA5 EOF. The individual ensemble members display independent NAO variability and average to zero in the ensemble mean. Figure \ref{fig:10}a and c display the corresponding anomaly regression patterns. The MDv0.9 anomaly regression pattern is calculated by pooling all ensemble members and fitting one joint regression. The spatial patterns match well, indicating that MDv0.9 displays a robust NAO-like mode of variability. 

The first EOF of the MDv0.9 ensemble Z500 anomaly (not shown) is also a meridional dipole, with negative loadings at the pole; we posit that this is a rotation of NAO variability, and use the ERA5-based EOF for comparison to assess physical realism. We use the MDv0.9 climatology-forced run for the comparison, controlling for interannual and longer-term variability that could be introduced by the historical oceanic forcings. ERA5 Z500 was detrended and anomalized prior to fitting the EOF. Although each MDv0.9 member exhibits NAO-like variability, its magnitude is dampened compared to that of ERA5.

\section{Discussion}
Here, we present the architectural and training details of MDv0.9 and perform an initial analysis of bias, forced responses, ENSO teleconnections, and atmospheric internal variability, as shown in a series of 20-member forced ensemble experiments. MDv0.9 displays significant biases in the atmospheric general circulation and spatial distributions of precipitation and temperature, as well as generally reduced-strength ENSO teleconnections and damped NAO-like atmospheric variability. In future versions of this model, the land surface and topography must be addressed and better represented. We speculate that increased vertical resolution could help resolve other internal atmospheric variability, such as the QBO. Additionally, we must further explore our model's response to initial conditions and refine our experimental protocol to allow sufficient ``spin-up" time. Despite all of these issues, we remain enthusiastic; MDv0.9 is extremely data-restricted ($N=372$) yet still performed 46.25-year auto-regressive ensemble experiments with stable results, which we deem reasonable given the circumstances. 
\newpage

\bibliographystyle{plainnat}
\bibliography{references}

@article{DBLP:journals/corr/abs-2112-10752,
  author       = {Robin Rombach and
                  Andreas Blattmann and
                  Dominik Lorenz and
                  Patrick Esser and
                  Bj{\"{o}}rn Ommer},
  title        = {High-Resolution Image Synthesis with Latent Diffusion Models},
  journal      = {CoRR},
  volume       = {abs/2112.10752},
  year         = {2021},
  url          = {https://arxiv.org/abs/2112.10752},
  eprinttype    = {arXiv},
  eprint       = {2112.10752},
  bibsource    = {dblp computer science bibliography, https://dblp.org}
}

@article{DBLP:journals/corr/abs-2102-09672,
  author       = {Alex Nichol and
                  Prafulla Dhariwal},
  title        = {Improved Denoising Diffusion Probabilistic Models},
  journal      = {CoRR},
  volume       = {abs/2102.09672},
  year         = {2021},
  url          = {https://arxiv.org/abs/2102.09672},
  eprinttype    = {arXiv},
  eprint       = {2102.09672},
  bibsource    = {dblp computer science bibliography, https://dblp.org}
}

@article{DBLP:journals/corr/abs-2006-11239,
  author       = {Jonathan Ho and
                  Ajay Jain and
                  Pieter Abbeel},
  title        = {Denoising Diffusion Probabilistic Models},
  journal      = {CoRR},
  volume       = {abs/2006.11239},
  year         = {2020},
  url          = {https://arxiv.org/abs/2006.11239},
  eprinttype    = {arXiv},
  eprint       = {2006.11239},
  bibsource    = {dblp computer science bibliography, https://dblp.org}
}

@article{diederik_introduction_2019,
	title = {An {Introduction} to {Variational} {Autoencoders}},
	volume = {12},
	issn = {1935-8237},
	url = {https://doi.org/10.1561/2200000056},
	doi = {10.1561/2200000056},
	number = {4},
	journal = {Foundations and Trends in Machine Learning},
	author = {Diederik, P. Kingma and Max, Welling},
	month = nov,
	year = {2019},
	note = {\_eprint: https://www.emerald.com/ftmal/article-pdf/12/4/307/11160827/2200000056en.pdf},
	pages = {307--392},
}

@misc{bonev2023sphericalfourierneuraloperators,
      title={Spherical Fourier Neural Operators: Learning Stable Dynamics on the Sphere}, 
      author={Boris Bonev and Thorsten Kurth and Christian Hundt and Jaideep Pathak and Maximilian Baust and Karthik Kashinath and Anima Anandkumar},
      year={2023},
      eprint={2306.03838},
      archivePrefix={arXiv},
      primaryClass={cs.LG},
      url={https://arxiv.org/abs/2306.03838}, 
}

@article{FiLM,
  author       = {Ethan Perez and
                  Florian Strub and
                  Harm de Vries and
                  Vincent Dumoulin and
                  Aaron C. Courville},
  title        = {FiLM: Visual Reasoning with a General Conditioning Layer},
  journal      = {CoRR},
  volume       = {abs/1709.07871},
  year         = {2017},
  url          = {http://arxiv.org/abs/1709.07871},
  eprinttype    = {arXiv},
  eprint       = {1709.07871},
  bibsource    = {dblp computer science bibliography, https://dblp.org}
}

@inproceedings{cvae,
	title = {Learning {Structured} {Output} {Representation} using {Deep} {Conditional} {Generative} {Models}},
	volume = {28},
	url = {https://proceedings.neurips.cc/paper_files/paper/2015/file/8d55a249e6baa5c06772297520da2051-Paper.pdf},
	booktitle = {Advances in {Neural} {Information} {Processing} {Systems}},
	publisher = {Curran Associates, Inc.},
	author = {Sohn, Kihyuk and Lee, Honglak and Yan, Xinchen},
	editor = {Cortes, C. and Lawrence, N. and Lee, D. and Sugiyama, M. and Garnett, R.},
	year = {2015},
}

@article{kochkov_neural_2024,
	title = {Neural general circulation models for weather and climate},
	volume = {632},
	issn = {1476-4687},
	url = {https://doi.org/10.1038/s41586-024-07744-y},
	doi = {10.1038/s41586-024-07744-y},
	number = {8027},
	journal = {Nature},
	author = {Kochkov, Dmitrii and Yuval, Janni and Langmore, Ian and Norgaard, Peter and Smith, Jamie and Mooers, Griffin and Klöwer, Milan and Lottes, James and Rasp, Stephan and Düben, Peter and Hatfield, Sam and Battaglia, Peter and Sanchez-Gonzalez, Alvaro and Willson, Matthew and Brenner, Michael P. and Hoyer, Stephan},
	month = aug,
	year = {2024},
	pages = {1060--1066},
}

@misc{lam2023graphcastlearningskillfulmediumrange,
      title={GraphCast: Learning skillful medium-range global weather forecasting}, 
      author={Remi Lam and Alvaro Sanchez-Gonzalez and Matthew Willson and Peter Wirnsberger and Meire Fortunato and Ferran Alet and Suman Ravuri and Timo Ewalds and Zach Eaton-Rosen and Weihua Hu and Alexander Merose and Stephan Hoyer and George Holland and Oriol Vinyals and Jacklynn Stott and Alexander Pritzel and Shakir Mohamed and Peter Battaglia},
      year={2023},
      eprint={2212.12794},
      archivePrefix={arXiv},
      primaryClass={cs.LG},
      url={https://arxiv.org/abs/2212.12794}, 
}

@misc{zhuang_pangeo-dataxesmf_2025,
	title = {pangeo-data/{xESMF}: v0.8.10},
	url = {https://doi.org/10.5281/zenodo.15304267},
	doi = {10.5281/zenodo.15304267},
	publisher = {Zenodo},
	author = {Zhuang, Jiawei and dussin, raphael and Huard, David and Bourgault, Pascal and Banihirwe, Anderson and Raynaud, Stephane and Malevich, Brewster and Schupfner, Martin and {Filipe} and Gauthier, Charles and Levang, Sam and Jüling, André and Almansi, Mattia and {RichardScottOZ} and {RondeauG} and Rasp, Stephan and Smith, Trevor James and Mares, Ben and Stachelek, Jemma and Plough, Matthew and {Pierre} and Bell, Ray and Caneill, Romain and Li, Xianxiang},
	month = apr,
	year = {2025},
}

@misc{lang_aifs_2024,
	title = {{AIFS} – {ECMWF}'s data-driven forecasting system},
	url = {https://arxiv.org/abs/2406.01465},
	author = {Lang, Simon and Alexe, Mihai and Chantry, Matthew and Dramsch, Jesper and Pinault, Florian and Raoult, Baudouin and Clare, Mariana C. A. and Lessig, Christian and Maier-Gerber, Michael and Magnusson, Linus and Bouallègue, Zied Ben and Nemesio, Ana Prieto and Dueben, Peter D. and Brown, Andrew and Pappenberger, Florian and Rabier, Florence},
	year = {2024},
	note = {\_eprint: 2406.01465},
}

@article{salimans2022progressive,
  author       = {Salimans, Tim and Ho, Jonathan},
  title        = {Progressive Distillation for Fast Sampling of Diffusion Models},
  journaltitle = {arXiv preprint arXiv:2202.00512},
  year         = {2022},
  eprint       = {2202.00512},
  eprinttype   = {arXiv},
  url          = {https://arxiv.org/abs/2202.00512}
}

@misc{brenowitz_climate_2025,
	title = {Climate in a {Bottle}: {Towards} a {Generative} {Foundation} {Model} for the {Kilometer}-{Scale} {Global} {Atmosphere}},
	url = {https://arxiv.org/abs/2505.06474},
	author = {Brenowitz, Noah D. and Ge, Tao and Subramaniam, Akshay and Manshausen, Peter and Gupta, Aayush and Hall, David M. and Mardani, Morteza and Vahdat, Arash and Kashinath, Karthik and Pritchard, Michael S.},
	year = {2025},
	note = {\_eprint: 2505.06474},
}

@misc{pathak_fourcastnet_2022,
	title = {{FourCastNet}: {A} {Global} {Data}-driven {High}-resolution {Weather} {Model} using {Adaptive} {Fourier} {Neural} {Operators}},
	url = {https://arxiv.org/abs/2202.11214},
	author = {Pathak, Jaideep and Subramanian, Shashank and Harrington, Peter and Raja, Sanjeev and Chattopadhyay, Ashesh and Mardani, Morteza and Kurth, Thorsten and Hall, David and Li, Zongyi and Azizzadenesheli, Kamyar and Hassanzadeh, Pedram and Kashinath, Karthik and Anandkumar, Animashree},
	year = {2022},
	note = {\_eprint: 2202.11214},
}

@misc{cachay_probabilistic_2024,
	title = {Probabilistic {Emulation} of a {Global} {Climate} {Model} with {Spherical} {DYffusion}},
	url = {https://arxiv.org/abs/2406.14798},
	author = {Cachay, Salva Rühling and Henn, Brian and Watt-Meyer, Oliver and Bretherton, Christopher S. and Yu, Rose},
	year = {2024},
	note = {\_eprint: 2406.14798},
}

@article{price_probabilistic_2025,
	title = {Probabilistic weather forecasting with machine learning},
	volume = {637},
	issn = {1476-4687},
	url = {https://doi.org/10.1038/s41586-024-08252-9},
	doi = {10.1038/s41586-024-08252-9},
	number = {8044},
	journal = {Nature},
	author = {Price, Ilan and Sanchez-Gonzalez, Alvaro and Alet, Ferran and Andersson, Tom R. and El-Kadi, Andrew and Masters, Dominic and Ewalds, Timo and Stott, Jacklynn and Mohamed, Shakir and Battaglia, Peter and Lam, Remi and Willson, Matthew},
	month = jan,
	year = {2025},
	pages = {84--90},
}

@article{hersbach_era5_2020,
	title = {The {ERA5} global reanalysis},
	volume = {146},
	url = {https://rmets.onlinelibrary.wiley.com/doi/abs/10.1002/qj.3803},
	doi = {https://doi.org/10.1002/qj.3803},
	number = {730},
	journal = {Quarterly Journal of the Royal Meteorological Society},
	author = {Hersbach, Hans and Bell, Bill and Berrisford, Paul and Hirahara, Shoji and Horányi, András and Muñoz-Sabater, Joaquín and Nicolas, Julien and Peubey, Carole and Radu, Raluca and Schepers, Dinand and Simmons, Adrian and Soci, Cornel and Abdalla, Saleh and Abellan, Xavier and Balsamo, Gianpaolo and Bechtold, Peter and Biavati, Gionata and Bidlot, Jean and Bonavita, Massimo and De Chiara, Giovanna and Dahlgren, Per and Dee, Dick and Diamantakis, Michail and Dragani, Rossana and Flemming, Johannes and Forbes, Richard and Fuentes, Manuel and Geer, Alan and Haimberger, Leo and Healy, Sean and Hogan, Robin J. and Hólm, Elías and Janisková, Marta and Keeley, Sarah and Laloyaux, Patrick and Lopez, Philippe and Lupu, Cristina and Radnoti, Gabor and de Rosnay, Patricia and Rozum, Iryna and Vamborg, Freja and Villaume, Sebastien and Thépaut, Jean-Noël},
	year = {2020},
	note = {\_eprint: https://rmets.onlinelibrary.wiley.com/doi/pdf/10.1002/qj.3803},
	pages = {1999--2049},
}

@misc{kingma2022autoencodingvariationalbayes,
      title={Auto-Encoding Variational Bayes}, 
      author={Diederik P Kingma and Max Welling},
      year={2022},
      eprint={1312.6114},
      archivePrefix={arXiv},
      primaryClass={stat.ML},
      url={https://arxiv.org/abs/1312.6114}, 
}

@misc{burgess2018understandingdisentanglingbetavae,
      title={Understanding disentangling in $\beta$-VAE}, 
      author={Christopher P. Burgess and Irina Higgins and Arka Pal and Loic Matthey and Nick Watters and Guillaume Desjardins and Alexander Lerchner},
      year={2018},
      eprint={1804.03599},
      archivePrefix={arXiv},
      primaryClass={stat.ML},
      url={https://arxiv.org/abs/1804.03599}, 
}
\end{document}